\pdfoutput=1
\documentclass{article}
\usepackage[T1]{fontenc}
\usepackage{times}

\usepackage[left=31mm,right=31mm,top=33mm,bottom=20mm]{geometry}
\usepackage{natbib}
\usepackage{booktabs}       
\usepackage{amsmath,amssymb,amsfonts}
\usepackage[thmmarks, amsmath, standard]{ntheorem}
\usepackage{bm}

\usepackage[lined,ruled]{algorithm2e}

\usepackage{enumitem}

\usepackage[usenames,dvipsnames]{xcolor}
\usepackage[colorlinks,
            linkcolor=blue,
            citecolor=blue,
            urlcolor=magenta,
            linktocpage,
            plainpages=false]{hyperref}

\bibpunct{(}{)}{;}{a}{,}{,}

\DeclareMathOperator*{\argmax}{arg\,max}

\DeclareMathOperator*{\E}{\mathsf{E}}

\newcommand{\phistar}{m}
\newcommand{\phimax}{\phi_{\max}}

\colorlet{leftcolor}{Cerulean}

\definecolor{rightcolor}{rgb}{0.9,0,0}

\newcommand{\morediscussion}[1]{} 

\renewtheorem{theorem}{Theorem}
\renewtheorem{lemma}{Lemma}
\renewtheorem{definition}{Definition}
\renewtheorem{corollary}{Corollary}

\title{An improved regret analysis for UCB-N and TS-N}

\author{Nishant A. Mehta \\ 
             University of Victoria \\
             \texttt{nmehta@uvic.ca}}

\begin{document}

\maketitle

\begin{abstract}
In the setting of stochastic online learning with undirected feedback graphs, \cite{lykouris2020feedback} previously analyzed the pseudo-regret of the upper confidence bound-based algorithm UCB-N and the Thompson Sampling-based algorithm TS-N. In this note, we show how to improve their pseudo-regret analysis. Our improvement involves refining a key lemma of the previous analysis, allowing a $\log(T)$ factor to be replaced by a factor $\log_2(\alpha) + 3$ for $\alpha$ the independence number of the feedback graph.
\end{abstract}

\section{Introduction}

This note concerns stochastic online learning with undirected feedback graphs, a sequential decision-making problem with a feedback level that can range from bandit feedback --- giving stochastic multi-armed bandits \citep{lai1985asymptotically,auer2002finite} --- to full-information feedback --- giving decision-theoretic online learning (DTOL)\footnote{Technically it is not quite DTOL as the learning algorithm must commit to a single arm in each round, although it may be do in a randomized way.} \citep{freund1997decision} under a stochastic i.i.d.~adversary.

In this problem setting, there is a finite set of arms $[K] = \{1, 2, \ldots, K\}$ and an undirected feedback graph $G = (V, E)$ with vertex set $V = [K]$ and a set of undirected edges $E \subseteq 2^V$ (with all self-loops included). The arms have an unknown joint reward distribution $P$ over $[0, 1]^K$, with each arm $j$'s marginal distribution $P_j$ having mean $\mu_j \in [0, 1]$. In each round $t$:
\begin{itemize}
\item A stochastic reward vector $X_t = (X_{t,a})_{a \in [K]}$ is drawn from $P$.
\item The learning algorithm pulls an arm $a_t \in [K]$ and collects reward $X_{t,a_t}$.
\item The learning algorithm observes the reward $X_{t,a}$ for all $a \in [K]$ such that $(a_t, a) \in E$.
\end{itemize}
The goal of the learning algorithm is to maximize its expected cumulative reward over $t$ rounds.

Without loss of generality, we index the arms so that $\mu_1 \geq \mu_2 \geq \ldots \geq \mu_K$. In the stochastic setting, our main interest is to bound the \emph{pseudo-regret}, defined as
\begin{align*}
\bar{R}_T := 
\max_{a \in [K]} \E \left[ \sum_{t=1}^T X_{t,a} -  \sum_{t=1}^T X_{t,a_t} \right] 
 = T \mu_1 - \E \left[ \sum_{t=1}^T X_{t,a_t} \right] .
\end{align*}
Letting $\Delta_a = \mu_1 - \mu_a$ for each $a \in [K]$, it is easy to show that the pseudo-regret is equal to 
\begin{align*}
\E \left[ \sum_{t=1}^T \Delta_{a_t} \right] .
\end{align*}

Recently, \citet[Theorems 6 and 12]{lykouris2020feedback} showed how both the upper confidence bound-style algorithm UCB-N and the Thompson Sampling-style algorithm TS-N obtain pseudo-regret of order at most
\begin{align}
\log(K T) \log (T) \max_{I \in \mathcal{I}(G)} \sum_{a \in I} \frac{1}{\Delta_a} , \label{eqn:original-regret}
\end{align}
where $\mathcal{I}(G)$ is the set of all independent sets of the graph $G$.

In this note, we will show how to improve the above result to one of order 
\begin{align}
\log(K T) \log_2(\alpha) \max_{I \in \mathcal{I}(G)} \sum_{a \in I} \frac{1}{\Delta_a} , \label{eqn:improved-regret}
\end{align}
where $\alpha$ is the independence number of $G$. To be clear, our analysis is still based upon the brilliant, layer-based analysis of \cite{lykouris2020feedback}; we simply refine one of their key lemmas (their Lemma 3) to obtain the improvement. 
In their work, \cite{lykouris2020feedback} asked the question of whether their extra $\log(T)$ factor, could be removed. While we have not entirely removed this factor, replacing it by $\log_2(\alpha)$ is arguably a great improvement. On the other hand, if one instead replaced $\log(T)$ by $\log(K)$, this might not be much of an improvement at all; indeed, in full-information settings, we often imagine that $K$ is exponential in $T$, meaning that $\log(T)$ may be \emph{preferable} to $\log(K)$. On the other hand, in such settings, we also have that $\alpha$ is very small (and $\log_2(\alpha)$ all the smaller). Yet, this begs the question of whether even the $\log_2(\alpha)$ factor is needed for UCB-N and TS-N. We conjecture that with the current, phase-based analysis, this factor is unavoidable, but leave open the possibility that a different analysis could remove this factor.

\section{Preliminaries} 

For each nonnegative integer $\phi$, define $G_\phi$ to be the subgraph induced by the vertices $a$ satisfying
\begin{align*}
2^{-\phi} < \Delta_a \leq 2^{-\phi + 1} .
\end{align*}
For some choices of $\phi$, the subgraph may have no vertices. We need only consider $\phi \leq \phimax$ for
\begin{align*}
\phimax := \min \left\{ \log(T), \left\lfloor \log_2 \frac{1}{\Delta_{\min}} \right\rfloor + 1 \right\} .  
\end{align*}

Let $L = 8 \log(2 T K / \delta)$ for $\delta = 1/T$. Then from the proof of Lemma 3 of \cite{lykouris2020feedback}, the main quantity to bound is
\begin{align}
\sum_{\phi=1}^{\phimax} \max_{I \in \mathcal{I}(G_\phi)} \sum_{a \in I} \frac{L}{2^{-2 \phi}} \cdot \Delta_a 
&\leq L \sum_{\phi=1}^{\phimax} \max_{I \in \mathcal{I}(G_\phi)} \sum_{a \in I} \frac{1}{2^{-2 \phi}} \cdot 2^{-\phi + 1} \nonumber \\
&\leq 2 L \sum_{\phi=1}^{\phimax} \max_{I \in \mathcal{I}(G_\phi)} \sum_{a \in I} 2^\phi . \label{eqn:start}
\end{align}

\cite{lykouris2020feedback} obtained the RHS above, except they considered the sum all the way up to $\phi = \lfloor \log(T) \rfloor$. They reasoned that there are at most $\log(T)$ values for $\phi$ that have contribution more than 1, and so the above is at most 1 plus
\begin{align*}
2 L \log(T) \max_\phi \max_{I \in \mathcal{I}(G_\phi)} \sum_{a \in I} 2^\phi
&\leq 4 L \log(T) \max_\phi \max_{I \in \mathcal{I}(G_\phi)} \sum_{a \in I} \frac{1}{\Delta_a} \\
&\leq 4 L \log(T) \max_{I \in \mathcal{I}(G)} \sum_{a \in I} \frac{1}{\Delta_a} .
\end{align*}

Via this reasoning, they obtained their Lemma 3, restated below for convenience.
\begin{lemma} \label{lemma:original}
Let $\Lambda_a^t$ be the highest layer arm $a$ is placed until time step $t$. Then
\begin{align*}
\sum_{t=1}^T \sum_{a \in [K]} 
    \Pr \left( a_t = a, \Lambda_a^t \leq \frac{L}{\Delta_a^2} \right) \Delta_a 
\leq 4 L \log(T)
        \max_{I \in \mathcal{I}(G)} \sum_{a \in I} \frac{1}{\Delta_a} 
        + 1 .
\end{align*}
\end{lemma}

\section{Improved result}

In this section, we show how to obtain the following refinement of Lemma~\ref{lemma:original} (Lemma 3 of \cite{lykouris2020feedback}):
\begin{lemma} \label{lemma:improved}
Let $\Lambda_a^t$ be the highest layer arm $a$ is placed until time step $t$. Then
\begin{align*}
\sum_{t=1}^T \sum_{a \in [K]} 
    \Pr \left( a_t = a, \Lambda_a^t \leq \frac{L}{\Delta_a^2} \right) \Delta_a 
\leq 4 L \left( \log_2(\alpha) + 3 \right)
        \max_{I \in \mathcal{I}(G)} \sum_{a \in I} \frac{1}{\Delta_a} 
        + 1 .
\end{align*}
\end{lemma}

Note that the $\log(T)$ factor has been replaced by $\log_2(\alpha) + 3$.

\begin{proof}[of Lemma~\ref{lemma:improved}]
Our departure point will be the summation in the RHS of \eqref{eqn:start}, rewritten as
\begin{align}
\sum_{\phi=1}^{\phimax} \max_{I \in \mathcal{I}(G_\phi)} \sum_{a \in I} 2^{\phi} . \label{eqn:start-2}
\end{align}

For each $\phi$, define $I_\phi := \argmax_{I \in \mathcal{I}(G_\phi)} \sum_{a \in I} 2^\phi$, and let $K_\phi := |I_\phi|$ be the corresponding cardinality. Using this notation, \eqref{eqn:start-2} may be re-expressed as
\begin{align}
\sum_{\phi=1}^{\phimax} K_\phi \cdot 2^{\phi} \label{eqn:start-3}
\end{align}

The subsequent analysis revolves around the following maximizing value of $\phi$:
\begin{align*}
\phistar := \argmax_{\phi \in \{1, 2, \ldots, \phimax \}} K_\phi \cdot 2^\phi .
\end{align*}
We will show that the sum \eqref{eqn:start-3} is essentially within a $\log_2(\alpha)$ multiplicative factor of $K_m \cdot 2^m$.

The first step is to decompose the summation \eqref{eqn:start-3} as
\begin{align*}
\sum_{\phi=1}^{\phimax} K_\phi \cdot 2^{\phi} 
= \textcolor{leftcolor}{\sum_{\phi=1}^{m-1} K_\phi \cdot 2^{\phi}} 
    + K_m \cdot 2^m 
    + \textcolor{rightcolor}{\sum_{m+1}^{\phimax} K_\phi \cdot 2^{\phi}} 
\end{align*}
We bound the RHS's second summation ($\phi > m$) and first summation $(\phi < m$) in turn.

\subsubsection*{\textcolor{rightcolor}{Sum over $\bm{\phi > m}$}}

Potentially overcounting, let us bound the objective of the following optimization problem:
\begin{equation*}
\begin{aligned}
& \underset{K_{\phistar + 1}, K_{\phistar + 2}, \ldots}{\text{maximize}}
& & \sum_{j=1}^\infty K_{\phistar + j} \cdot 2^{\phistar + j}  \\
& \text{subject to}
& & K_{\phistar + j} \cdot 2^{\phistar + j} \leq K_{\phistar} \cdot 2^{\phistar} , \; j = 1, 2, \ldots .
\end{aligned}
\end{equation*}
The constraints, arising from the maximizing property of $m$, trivially may be rewritten as
\begin{align*}
K_{\phistar + j} \leq K_{\phistar} \cdot 2^{-j} , \; j = 1, 2, \ldots .
\end{align*}

Clearly, for any $j$ such that $K_{m + j}$ only has zero as the sole feasible integer value, the associated term $K_{m+j} \cdot 2^{m+j}$ can be ignored in the objective. Therefore, let us find the largest $j$ such that $K_{\phistar} \cdot 2^{-j} \geq 1$, which is $j_1 := \lfloor \log_2(K_{\phistar}) \rfloor$. From the maximizing property of $m$, the optimal value of the above problem is therefore at most $j_1 \cdot K_m \cdot 2^m$.

\subsubsection*{\textcolor{leftcolor}{Sum over $\bm{\phi < m}$}}

Again potentially overcounting, we will now bound the objective of the below problem:
\begin{equation*}
\begin{aligned}
& \underset{K_{\phistar - 1}, K_{\phistar - 2}, \ldots}{\text{maximize}}
& & \sum_{j=1}^\infty K_{\phistar - j} \cdot 2^{\phistar - j}  \\
& \text{subject to}
& & K_{\phistar - j} \cdot 2^{\phistar - j} \leq K_{\phistar} \cdot 2^{\phistar} , \; j = 1, 2, \ldots .
\end{aligned}
\end{equation*}

We first rewrite the constraints as
\begin{align*}
K_{\phistar - j} \leq K_{\phistar} \cdot 2^j , \; j = 1, 2, \ldots .
\end{align*}

Now, in order to maximize the summation, for as many values of $j$ as possible we should set $K_{m-j} = K_m \cdot 2^j$. However, since each $K_{m-j}$ is the size of an independent set of a subgraph of $G$, we must have that all such $K_{m-j} \leq \alpha$. 
Therefore, let us find the smallest $j$ such that $K_{\phistar} \cdot 2^j \geq \alpha$, which is $j_2 = \left\lceil \log_2 \left( \frac{\alpha}{K_{\phistar}} \right) \right\rceil$. For $j = 1, 2, \ldots, j_2$, we simply upper bound $K_{m-j} \cdot 2^j$ by the maximum possible value $K_m \cdot 2^j$. However, as $j$ increases beyond $j_2$, we have that $K_{\phistar - j}$ can no longer grow (since $\alpha$ is the largest possible value), and so $K^{\phistar - j} \cdot 2^{\phistar - j}$ geometrically decreases. Consequently, cumulatively over all such $j$ beyond $j_2$, the contribution to the summation is at most a single term $K_m \cdot 2^m$. Hence, the optimal value of the above problem is at most $(j_2 + 1) \cdot K_m \cdot 2^m$.

\subsubsection*{Putting everything together}

Putting together the two pieces above and accounting for the term due to $m$ itself, it holds that
\begin{align*}
\sum_{\phi=1}^{\phimax} K_\phi \cdot 2^{\phi} 
&\leq (j_1 + j_2 + 2) \cdot K_m \cdot 2^m \\
&= \left( 
             \lfloor \log_2(K_{\phistar}) \rfloor 
             + \left\lceil \log_2 \left( \frac{\alpha}{K_{\phistar}} \right) \right\rceil
             + 2
         \right)
         \cdot K_{\phistar} \cdot 2^{\phistar} \\
&\leq \left( \log_2(\alpha) + 3 \right)
         \cdot K_{\phistar} \cdot 2^{\phistar} \\
&= \left( \log_2(\alpha) + 3 \right)
         \max_\phi \max_{I \in \mathcal{I}(G_\phi)} \sum_{a \in I} 2^\phi \\
&\leq 2 \left( \log_2(\alpha) + 3 \right)
         \max_\phi \max_{I \in \mathcal{I}(G_\phi)} \sum_{a \in I} \frac{1}{\Delta_a} \\
&\leq 2 \left( \log_2(\alpha) + 3 \right)
         \max_{I \in \mathcal{I}(G)} \sum_{a \in I} \frac{1}{\Delta_a} .
\end{align*}
\end{proof}

\section{Discussion}

The improvement to Lemma 3 of \cite{lykouris2020feedback} given by our Lemma~\ref{lemma:improved} leads to the same improvement in their result for UCB-N and TS-N (their Theorems 6 and 12 respectively), as well as replacing the $\log(T)$ in their gap-independent bounds Corollaries 7 and 13 by a term of order $\log_2(\alpha)$. 
For concreteness, we stated the improved problem-dependent and problem-independent regret bounds for UCB-N; it is straightforward to fill in the improved regret bounds for TS-N.

\begin{theorem}
With the setting $\delta = \frac{1}{T}$, the pseudo-regret of the UCB-N algorithm (Algorithm 2 of \cite{lykouris2020feedback}) can be bounded as
\begin{align*}
\bar{R}_T \leq 8 \log(2 K T^2) \left( \log_2(\alpha) + 3 \right) \max_{I \in \mathcal{I}(G)} \sum_{a \in I} \frac{1}{\Delta_a} + 2 .
\end{align*}
\end{theorem}

\begin{corollary}
The expected regret of UCB-N is bounded by
\begin{align*}
2 + 4 \sqrt{2 \alpha T \log(2 K T^2) \left(\log_2(\alpha) + 3 \right)} .
\end{align*}
\end{corollary}

\subsection*{Acknowledgements}
Many thanks to Bingshan Hu for putting up with my drawings of doubling rectangles back in November 2020, which is when the improvement herein was worked out. To the benefit of the reader, this note formalizes the ``rectangles'' argument, whose geometric visualization is sadly absent from the present version of this note. Also, thanks to Dirk van der Hoeven for continuing to encourage me to publish this note.

\bibliography{rectangles_reloaded}

\begin{thebibliography}{4}
\providecommand{\natexlab}[1]{#1}
\providecommand{\url}[1]{\texttt{#1}}
\expandafter\ifx\csname urlstyle\endcsname\relax
  \providecommand{\doi}[1]{doi: #1}\else
  \providecommand{\doi}{doi: \begingroup \urlstyle{rm}\Url}\fi

\bibitem[Auer et~al.(2002)Auer, Cesa-Bianchi, and Fischer]{auer2002finite}
Peter Auer, Nicolo Cesa-Bianchi, and Paul Fischer.
\newblock Finite-time analysis of the multiarmed bandit problem.
\newblock \emph{Machine learning}, 47:\penalty0 235--256, 2002.

\bibitem[Freund and Schapire(1997)]{freund1997decision}
Yoav Freund and Robert~E Schapire.
\newblock A decision-theoretic generalization of on-line learning and an
  application to boosting.
\newblock \emph{Journal of computer and system sciences}, 55\penalty0
  (1):\penalty0 119--139, 1997.

\bibitem[Lai et~al.(1985)Lai, Robbins, et~al.]{lai1985asymptotically}
Tze~Leung Lai, Herbert Robbins, et~al.
\newblock Asymptotically efficient adaptive allocation rules.
\newblock \emph{Advances in applied mathematics}, 6\penalty0 (1):\penalty0
  4--22, 1985.

\bibitem[Lykouris et~al.(2020)Lykouris, Tardos, and Wali]{lykouris2020feedback}
Thodoris Lykouris, Eva Tardos, and Drishti Wali.
\newblock Feedback graph regret bounds for {T}hompson {S}ampling and {UCB}.
\newblock In \emph{Algorithmic Learning Theory}, pages 592--614. PMLR, 2020.

\end{thebibliography}

\end{document}